%% file: main.tex
\documentclass[10pt,twocolumn,letterpaper]{article}

\usepackage[accepted]{cvpr}

\input{preamble}
\definecolor{cvprblue}{rgb}{0.21,0.49,0.74}
\usepackage[pagebackref,breaklinks,colorlinks,allcolors=cvprblue]{hyperref}

\def\paperID{37207}
\def\confName{CVPR}
\def\confYear{2026}

\title{Learning with Adaptive Prototype Manifolds for Out-of-Distribution Detection}

\author{
    Ningkang Peng, Jiutao Zhou, Yuhao Zhang, Qianfeng Yu, Linjing Qian \\
    Nanjing Normal University \\
    {\tt\small \{nkpeng, 242212052, yuhaozhang, qfyu, ljqian\}@nnu.edu.cn}
\and
    Xiaoqian Peng \\
    Nanjing University of Chinese Medicine \\
    {\tt\small 202411148@njucm.edu.cn}
\and
    Tingyu Lu$^*$ \\
    Tohoku University\\
    {\tt\small tingyu.lu.e7@tohoku.ac.jp}
\and    
    Yi Chen$^*$, Yanhui Gu\thanks{Corresponding authors.} \\
    Nanjing Normal University \\
    {\tt\small \{cs\_chenyi, gu\}@njnu.edu.cn}
}

\begin{document}
\maketitle
\input{sec/0_abstract}
\input{sec/1_intro}
\input{sec/2_Related}
\input{sec/3_Method}
\input{sec/4_Experimental}

\input{sec/5_Conclusion}
 {
     \small
     \bibliographystyle{ieeenat_fullname}
     \bibliography{main}
 }


\end{document}

%% file: preamble.tex
\usepackage{algorithm}
\usepackage{algorithmic}
\usepackage{multirow}
\usepackage{amsmath}
\usepackage{graphicx}
\usepackage{booktabs} 
\usepackage{siunitx}
\usepackage{caption}
\usepackage{subcaption}



%% file: sec/0_abstract.tex
\begin{abstract}
Out-of-distribution (OOD) detection is a critical task for the safe deployment of machine learning models in the real world. Existing prototype-based representation learning methods have demonstrated exceptional performance. Specifically, we identify two fundamental flaws that universally constrain these methods: the Static Homogeneity Assumption (fixed representational resources for all classes) and the Learning-Inference Disconnect (discarding rich prototype quality knowledge at inference). These flaws fundamentally limit the model's capacity and performance. To address these issues, we propose APEX (Adaptive Prototype for eXtensive OOD Detection), a novel OOD detection framework designed via a Two-Stage Repair process to optimize the learned feature manifold. APEX introduces two key innovations to address these respective flaws: (1) an Adaptive Prototype Manifold (APM), which leverages the Minimum Description Length (MDL) principle to automatically determine the optimal prototype complexity $K_c^*$ for each class, thereby fundamentally resolving prototype collision; and (2) a Posterior-Aware OOD Scoring (PAOS) mechanism, which quantifies prototype quality (cohesion and separation) to bridge the learning-inference disconnect. Comprehensive experiments on benchmarks such as CIFAR-100 validate the superiority of our method, where APEX achieves new state-of-the-art performance.
\end{abstract}

%% file: sec/1_intro.tex
\section{Introduction}

When deploying machine learning models in the open world, ensuring their safety and reliability in the face of unknown inputs is a critical first step\cite{c:25}. A model must not only make accurate predictions on known in-distribution (ID) data but also possess the ability to identify out-of-distribution (OOD) samples\cite{c:43} that it has not encountered during training. The task of OOD detection is dedicated to addressing this core challenge\cite{intro:early-method} and has become an indispensable component in building trustworthy AI systems\cite{intro:trustworthy-AI-systems,c:47}.

\begin{figure}[!ht]
    \centering \includegraphics[width=0.48\textwidth]{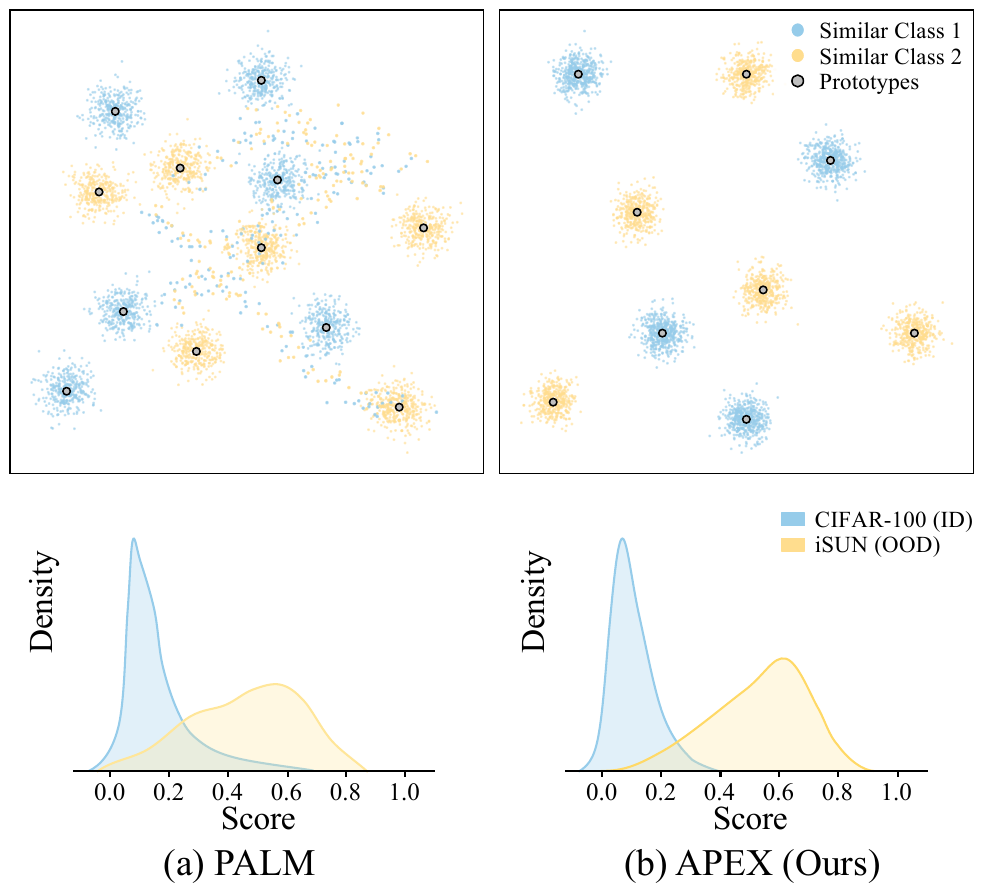} 
        \caption{
        The Pprototype collision defect and its Mitigation: While the baseline PALM model suffers from \textbf{prototype collision} due to a fixed number of prototypes, our APM model effectively resolves this issue by utilizing an adaptive optimal number of prototypes $K_c^*$, forming distinct and compact class clusters.
    }
    \label{fig:UMAP}
\end{figure}
 
To achieve effective OOD detection, the research community has explored various paradigms, including methods based on confidence scores, density estimation, and distance metrics. Among them, distance-based methods have shown immense potential. These approaches circumvent the overconfidence problem of traditional confidence scores on OOD samples, and are centered around learning a high-quality feature embedding space where ID data form compact clusters, making OOD samples identifiable by their relative distance.
Arguably, the efficacy of distance-based methods depends heavily on the quality of the feature embeddings. To this end, early works began to employ powerful contrastive learning objectives (e.g., SupCon\cite{c:27}) to learn more discriminative representations\cite{c:28,intro:discriminative-representations,c:40}. Building on this, prototypical learning has emerged as a prominent approach. This method further refines the characterization of the ID distribution by explicitly learning one or more representative prototypes for each ID class, achieving state-of-the-art performance. The success of these works strongly demonstrates that a well-structured representation space is key to solving the OOD problem. However, our in-depth analysis reveals that the frameworks of these methods still inherit two previously overlooked but fundamental theoretical flaws from this paradigm, which we systematically categorize as the Static Homogeneity Assumption and the Learning-Inference Disconnect.
The first flaw is the Static Homogeneity Assumption. Existing work tends to allocate homogeneous representational resources to all classes, such as a fixed number and equal importance of prototypes\cite{c:26}. This ignores the vast differences in intrinsic visual complexity between categories\cite{ignores-inter-class-variability,ignores-inter-class-variability2,visual-complexity-differences,visual-complexity-differences2}, for example, between an apple and a pride of lions, leading to suboptimal modeling efficiency. The second flaw is the Learning-Inference Disconnect. The rich posterior knowledge about the internal states of the model, learned during the training phase, is discarded at inference time in favor of a separate, information-simplified scoring function\cite{c:25,c:41,c:43}. This loss of information inevitably imposes a lower theoretical upper bound on performance\cite{c:45,EBMretain-information}.

To fundamentally address these problems, we first investigate the catastrophic structural failure caused by the Static Homogeneity Assumption: prototype collision. Our empirical analysis of the PALM model\cite{c:35} ($K=6$) reveals that a portion of prototypes severely collapse, forming discriminatory blind spots in the feature space. As shown in Figure \ref{fig:UMAP}, this collision occurs precisely between semantically similar but structurally distinct categories, such as leopard and tiger. This phenomenon proves that allocating homogeneous resources is suboptimal and dangerous, as the model is structurally incapable of distinguishing the known ID classes in these conflicted regions. Furthermore, the information loss inherent in the Learning-Inference Disconnect hinders our ability to leverage the rich knowledge gained during training (specifically, the quality information of each learned prototype) to compensate for these structural weaknesses during decision-making.
Guided by these structural insights, we propose APEX (Adaptive Prototype for eXtensive OOD Detection), a carefully designed OOD detection framework aimed at systematically repairing these two fundamental structural and informational problems. APEX achieves this repair through two orthogonal innovations. First, to fundamentally solve the collision problem, we introduce the Adaptive Prototype Manifold (APM). This mechanism, inspired by the Minimum Description Length (MDL) principle, adaptively allocates appropriate representational capacity for each class. Second, to bridge the information gap, we design the Posterior-Aware OOD Scoring (PAOS) mechanism. It fully integrates all posterior quality knowledge learned by the APM into the inference process, ensuring consistency between training and decision-making.

In summary, our main contributions are as follows:
\begin{itemize}
\item We are the first to identify and systematically articulate two fundamental flaws in sota prototype-based OOD detection methods, providing empirical evidence and visualization of catastrophic prototype collision, which is the core consequence of the static homogeneity assumption.
\item We propose the APEX framework and introduce its core component, the APM. This mechanism, for the first time, introduces the concept of heterogeneous representation learning to OOD detection, fundamentally solving the prototype collision problem caused by the static homogeneity assumption.
\item We design the PAOS function, which innovatively integrates the model's complete posterior quality knowledge into decision-making. It offers a new, principled paradigm to resolve the learning-inference disconnect by ensuring theoretical consistency between the training and inference objectives.
\item We conduct comprehensive experiments on multiple standard OOD detection benchmarks to validate the superiority of the APEX framework and demonstrate its significant performance improvements over existing state-of-the-art methods.
\end{itemize}

%% file: sec/2_Related.tex
\section{Related Work}

\paragraph{OOD Detection and Representation Learning}
OOD detection aims to identify test samples that differ significantly from the training data distribution, which is crucial for ensuring the safety and reliability of AI systems. Early works primarily focused on post-processing the model's outputs, giving rise to various scoring functions, such as those based on the Maximum Softmax Probability (MSP) \cite{c:25, c:43}, energy functions \cite{c:37, c:44, c:45}, or model gradients \cite{c:38}. Another major category, generative model-based methods, attempts to explicitly model the distribution of ID data, assuming that OOD samples will have a lower likelihood under the learned distribution \cite{c:39, c:46}. In recent years, distance-based methods have become a mainstream direction \cite{c:40, c:30,c:41,c:42}. The core idea of this paradigm is that within a feature space learned by a deep neural network, ID data form compact clusters, while OOD samples lie far from them \cite{c:26, c:41, c:27}. The success of these methods is highly dependent on the quality of the representation. Consequently, the research focus has gradually shifted towards learning a superior representation space for OOD detection. This evolution includes leveraging contrastive learning (e.g., SupCon) to enhance feature discriminability \cite{c:27, c:42}, synthesizing outliers to explicitly regularize the decision boundary (e.g., VOS, NPOS) \cite{c:29, c:30}, and embedding features onto a hypersphere for more fine-grained modeling of the ID distribution \cite{c:31,c:32,c:33,c:35}.

\paragraph{Prototypical Learning}
Among the various representation learning strategies, prototypical learning has emerged as a particularly effective method for fine-grained modeling \cite{c:33,c:34,c:62}. Its core idea is to learn one or more explicit, representative prototypes for each ID class \cite{c:30,c:34}. By encouraging samples during training to move closer to the prototypes of their own class and away from others, the model learns a more structured embedding space \cite{c:27,c:63}. For instance, CIDER learns a single prototype per class\cite{c:34}, while its significant successor, PALM learns a mixture of prototypes to better capture intra-class diversity\cite{c:35}. However, despite their considerable success, existing prototypical learning methods, whether for general classification or OOD detection, tend to adopt a generic, homogeneous design. For example, they typically assign a fixed number of prototypes and equal importance to all classes\cite{c:64,c:65}. This is the root of the two fundamental flaws we identified in the introduction.

%% file: sec/3_Method.tex
\begin{figure*}[!ht]
    \centering \includegraphics[width=0.95\textwidth]{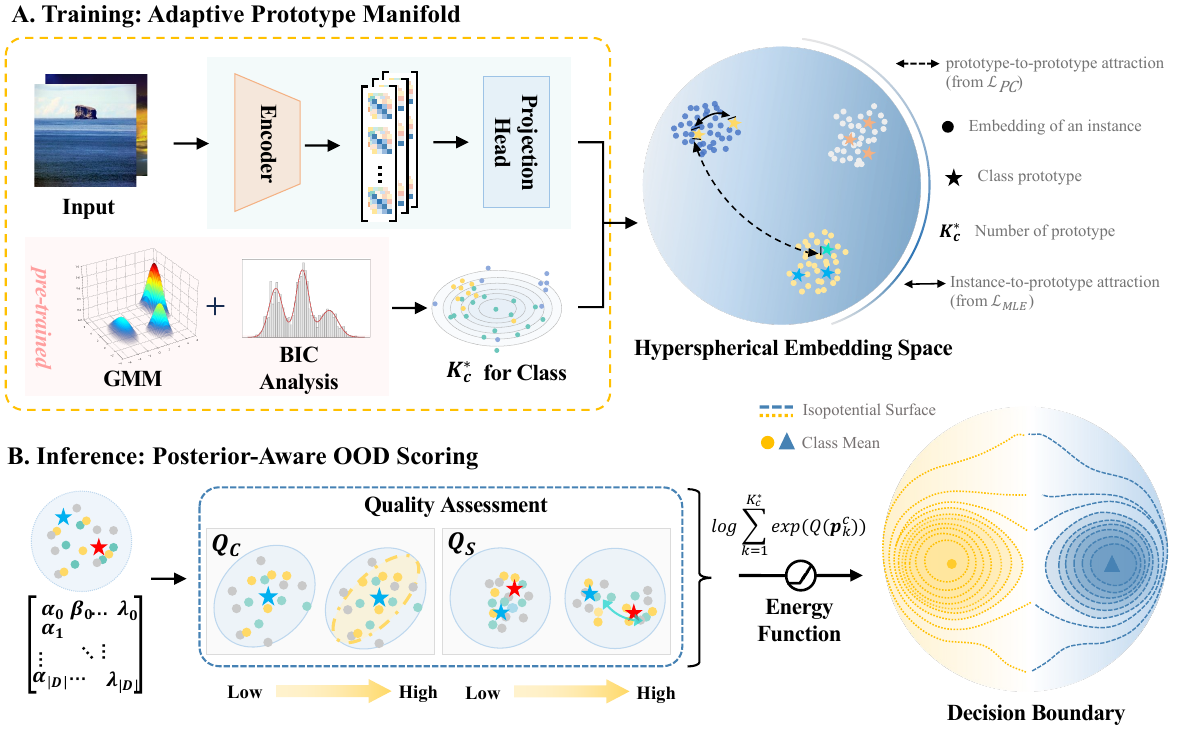} 
        \caption{
        The overall architecture of our Two-Stage Repair Framework . 
        \textbf{A. Training: Adaptive Prototype Manifold. }
        $\text{APM}$ determines the optimal prototype complexity $K_c^*$ for each class using Gaussian Mixture Model ($\text{GMM}$) and Bayesian Information Criterion ($\text{BIC}$) analysis, optimizing the hyperspherical embedding space for better structural separation. 
        \textbf{B. Inference: Posterior-Aware OOD Scoring. }
        $\text{PAOS}$ utilizes prototype quality information (left) to calibrate the final OOD scoring function (right), which is based on the Energy Function, achieving fine-grained decision boundary refinement.
    }
    \label{fig:framework}
\end{figure*}

\section{Methodology}
Our framework aims to address the core deficiencies in existing prototype learning methods through a Two-Stage Repair process, as shown in Figure \ref{fig:framework}. The first stage (\textbf{Structural Repair}) introduces APM, which determines the optimal prototype complexity $K_c^*$ for each class in a data-driven manner. The second stage (\textbf{Score Calibration}) introduces the PAOS mechanism, which utilizes prototype quality information to calibrate the OOD scoring function. This section will detail the technical aspects of these two stages.

\subsection{Adaptive Prototype Manifold}

\subsubsection{Determining the Optimal Prototype Complexity}
To resolve issues caused by the static homogeneity assumption, we first determine the optimal prototype complexity $K_c^*$ for each class $c$. The core idea is to construct a flexible and precise probabilistic model $p(\mathbf{z}\mid y)$, adjusting its representational capacity according to the intrinsic complexity of each class. This process follows the \text{MDL} principle\cite{MDL}, implemented via the \text{BIC}\cite{BIC}, and is executed once before training.
We utilize a powerful pre-trained DINOv2 model to extract high-quality feature vectors for all samples in the ID training set\cite{dino}. For the feature set of each class $c$, we fit a \text{GMM}, and use the \text{BIC} to determine the optimal number of clusters, which serves as our optimal prototype number $K_c^*$:
\begin{equation}
K_c^* = \underset{k' \in \mathcal{K}}{\text{argmin}} \left( N(k', D) \cdot \ln(n_c) - 2 \ln(\hat{L}_c(k')) \right),
\label{eq:bic}
\end{equation}
where $k'$ is the candidate number of prototypes, $D$ is the feature dimension, $n_c$ is the number of samples in class $c$, and $\hat{L}_c(k')$ is the corresponding maximum likelihood. $N(k', D)$ is the total number of free parameters required for the GMM with $k'$ components in $D$ dimensions.

\subsubsection{Hierarchical Probabilistic Model}
We adopt a hierarchical probabilistic model and training objective similar to PALM, but critically, we use the adaptive $K_c^*$ instead of a fixed $K$. The model assumes the class-conditional probability density $p(\mathbf{z}_i|y_i=c)$ for a sample $\mathbf{z}_i$ is a mixture of von Mises-Fisher (vMF) distributions parameterized by the $K_c^*$ prototypes $\{\mathbf{p}_j^c\}_{j=1}^{K_c^*}$ of that class:
\begin{equation}
\begin{split} 
p(\mathbf{z}_i|y_i=c) &= \sum_{j=1}^{K_c^*} w_{i,j}^c f_{vMF}(\mathbf{z}_i; \mathbf{p}_j^c, \kappa) \\ 
&= C_D(\kappa) \sum_{j=1}^{K_c^*} w_{i,j}^c \exp(\kappa (\mathbf{p}_j^c)^T \mathbf{z}_i),
\end{split}
\label{eq:vmf_mixture}
\end{equation}
where $w_{i,j}^c$ is the soft assignment weight of sample $\mathbf{z}_i$ to prototype $\mathbf{p}_j^c$ (see Section~\ref{sec:soft_assignment}), and $\kappa$ is the concentration parameter of the vMF distribution.
Based on this model, the posterior probability of sample $\mathbf{z}_i$ belonging to its true class $y_i$ is:
\begin{equation}
P(y_i|\mathbf{z}_i) = \frac{\sum_{j=1}^{K_{y_i}^*} w_{i,j}^{y_i} \exp({\mathbf{p}_j^{y_i}}^T \mathbf{z}_i / \tau)}{\sum_{c'=1}^{C} \sum_{k=1}^{K_{c'}^*} w_{i,k}^{c'} \exp({\mathbf{p}_k^{c'}}^T \mathbf{z}_i / \tau)},
\label{eq:posterior_prob}
\end{equation}
where $\tau = 1/\kappa$ is the temperature parameter.

\subsubsection{Training Objective}
\begin{enumerate}
\item \textbf{Maximum Likelihood Estimation Loss ($\mathcal{L}_{MLE}$)}: Minimizes the average negative log-likelihood over all $N$ training samples:
\begin{equation}
\mathcal{L}_{MLE}(\mathcal{K}^*) = -\frac{1}{N} \sum_{i=1}^{N} \log P(y_i|\mathbf{z}_i),
\label{eq:LMLE}
\end{equation}
where $P(y_i|\mathbf{z}_i)$ is the posterior probability defined in Equation ~\ref{eq:posterior_prob}.
\item \textbf{Prototype Contrastive Loss ($\mathcal{L}_{PC}$)}: Enhances the cluster structure by pulling prototypes of the same class closer and pushing prototypes of different classes apart. We define the intra-class similarity term $D_{j}^{c} = \sum_{k=1, k \neq j}^{K_c^*} \exp(\mathbf{p}_j^c \mathbf{p}_k^c / \tau_p)$, and the inter-class similarity term $Z_{j}^{c} = \sum_{c' \neq c} \sum_{l=1}^{K_{c'}^*} \exp(\mathbf{p}_j^c \mathbf{p}_l^{c'} / \tau_p)$, for the $j$-th prototype of class $c$. The $\mathcal{L}_{PC}$ is expressed as:
  	\begin{equation}
  	\mathcal{L}_{PC}(\mathcal{K}^*) = -\frac{1}{M} \sum_{c=1}^{C} \sum_{j=1}^{K_c^*} \log \left( \frac{D_{j}^{c}}{Z_{j}^{c}} \right),
  	\label{eq:LPC}
  	\end{equation}
where $M = \sum K_c^*$, the term $\mathbf{p}_j^c \mathbf{p}_k^c$ denotes the cosine similarity between prototypes, and $\tau_p$ is a temperature parameter.
\end{enumerate}

The final total training loss is:
\begin{equation}
\mathcal{L}_{Train} = \mathcal{L}_{MLE}(\mathcal{K}^*) + \lambda \mathcal{L}_{PC}(\mathcal{K}^*).
\label{eq:LTrain}
\end{equation}
By minimizing $\mathcal{L}_{Train}$, we train the encoder $f_\theta$ and the projection head $g_\phi$ to learn a discriminative feature space with an adaptive manifold structure.

\subsection{Posterior-Aware OOD Scoring}
\subsubsection{Prototype Quality Assessment}
To bridge the learning-inference disconnect, we need to dynamically capture prototype quality during the training process. We define two core metrics:

\begin{itemize}
\item \textbf{Cohesion Score ($Q_C$)}: Measures the representativeness of prototype $\mathbf{p}_k^c$ for its own class $c$. It calculates the average cosine similarity between the prototype and its assigned sample set $S_k^c$:
    \begin{equation}
    Q_C(\mathbf{p}_k^c) = \text{mean}_{\mathbf{z}_i \in S_k^c} \left( \frac{(\mathbf{p}_k^c)^T \mathbf{z}_i}{||\mathbf{p}_k^c|| ||\mathbf{z}_i||} \right).
    \label{eq:QC}
    \end{equation}
\item \textbf{Separation Score ($Q_S$)}: Measures the uniqueness and inter-class discriminability of prototype $\mathbf{p}_k^c$. It calculates the cosine distance to the nearest prototype $\mathbf{p}_j^{c'}$ from a different class ($c' \neq c$):
    \begin{equation}
    Q_S(\mathbf{p}_k^c) = 1 - \max_{c' \neq c, \forall j \in \{1..K_{c'}^*\}} \left( \frac{(\mathbf{p}_k^c)^T \mathbf{p}_j^{c'}}{||\mathbf{p}_k^c|| ||\mathbf{p}_j^{c'}||} \right).
    \label{eq:QS}
    \end{equation}
\end{itemize}

We combine these two metrics into a final prototype quality score ($Q$):
\begin{equation}
Q(\mathbf{p}_k^c) = Q_C(\mathbf{p}_k^c) + Q_S(\mathbf{p}_k^c).
\label{eq:quality_score}
\end{equation}

\subsubsection{Category Energy}
\label{sec:category_energy}

To obtain an overall confidence for each class $c$, we aggregate the quality scores $Q(\mathbf{p}_k^c)$ of all its $K_c^*$ prototypes (defined in Equation~\ref{eq:quality_score}).
Inspired by EBMs, we frame this aggregation problem from a statistical mechanics perspective.
We define an energy for each prototype, where high quality corresponds to a low-energy state:
\begin{equation}
    E_q(\mathbf{p}_k^c) = -Q(\mathbf{p}_k^c).
\end{equation}
Following the principles of EBMs, we define a Gibbs distribution $p_c(k)$ over the set of $K_c^*$ prototypes within class $c$:
\begin{equation}
\begin{split}
    p_c(k) &= \frac{\exp(-E_q(\mathbf{p}_k^c) / \tau_q)}{\sum_{k'=1}^{K_c^*} \exp(-E_q(\mathbf{p}_{k'}^c) / \tau_q)} \\&=\frac{\exp(Q(\mathbf{p}_k^c) / \tau_q)}{\sum_{k'=1}^{K_c^*} \exp(Q(\mathbf{p}_{k'}^c) / \tau_q)},
\end{split}
\label{eq:gibbs_dist}
\end{equation}
where $\tau_q$ is a temperature hyperparameter. This formulation explicitly defines the probability of selecting a prototype $k$ based on its relative quality $Q(\mathbf{p}_k^c)$ within the class $c$.
In this context, the negative free energy $\mathcal{F}_q(c)$ is used to quantify the overall confidence of the class system. The free energy is directly related to the partition function $Z_c$ via $\mathcal{F}_q(c) = -\tau_q \log Z_c$.
A class's overall confidence $Conf(c)$, is then defined as the negative of the free energy:
\begin{equation}
\begin{split}
    Conf(c) &= - \mathcal{F}_q(c) 
    = - \left( -\tau_q \log Z_c \right) \\
    &= \tau_q \log \left( \sum_{k=1}^{K_c^*} \exp \left( \frac{Q(\mathbf{p}_k^c)}{\tau_q} \right) \right).
\end{split}
\label{eq:category_confidence}
\end{equation}
During inference, to address the learning-inference gap, we seamlessly integrate this principled, post-hoc class confidence $Conf(c)$ into our OOD scoring function.
We use the class confidence $Conf(c)$ to calibrate the class-wise Mahalanobis distance. High confidence classes (indicating the model's judgment in that region is more reliable) should have a stronger pull on nearby samples (i.e., the effective distance should be smaller), and vice versa. We adopt the following calibration formula:
\begin{equation}
S_{PAOS}(\mathbf{h}) = \min_{c \in \{1, ..., C\}} \left( \frac{(\mathbf{h} - \boldsymbol{\mu}_c)^T \Sigma^{-1} (\mathbf{h} - \boldsymbol{\mu}_c)}{1 + \alpha \cdot Conf(c)} \right),
\label{eq:qc_maha_score}
\end{equation}
where $\mathbf{h}$ is the input sample feature from the backbone network; $\boldsymbol{\mu}_c$ is the mean vector of class $c$; $\Sigma^{-1}$ is the inverse of the global covariance matrix. $\alpha$ is a hyperparameter controlling the calibration strength. The final OOD score $S_{PAOS}(\mathbf{h})$ is the minimum of all calibrated distances. The lower the score, the more likely the sample is to be ID.

\subsection{Soft Assignment and Parameter Update}
\label{sec:soft_assignment}
To implement the aforementioned training objectives and quality assessment, we need to define prototype assignment and update mechanisms.

\subsubsection{Soft Assignment Weights}
During training, the soft assignment weights $w_{i,k}^c$ between a sample $\mathbf{z}_i$ and the prototypes $\{\mathbf{p}_k^c\}_{j=1}^{K_c^*}$ of its class $c$ are not learned parameters. Instead, they are dynamically computed for each batch using a method based on Entropy-Regularized Optimal Transport. For the sample embedding matrix $Z^c$ and prototype matrix $P^c$ of class $c$ within a batch, the weight matrix $W^c$ (whose columns are $w_{i,k}^c$) is efficiently solved using the Sinkhorn-Knopp algorithm:
\begin{equation}
W^{c} = \text{diag}(\mathbf{u}) \exp\left(\frac{(P^{c})^T Z^{c}}{\epsilon}\right) \text{diag}(\mathbf{v}),
\end{equation}
where $\epsilon$ is the regularization strength, and $\mathbf{u}$ and $\mathbf{v}$ are the row and column scaling vectors, computed via Sinkhorn iterations to ensure $W^c$ satisfies the marginal constraints of the optimal transport problem. These weights $w_{i,k}^c$ are used in the calculation of $\mathcal{L}_{MLE}$ and the update of prototypes.

\subsubsection{Update of Prototypes }
The prototypes $\mathbf{p}_k^c$, serving as the central anchors for each sub-cluster, are smoothly updated using Exponential Moving Average (EMA)\cite{EMA} based on the weighted sum of sample embeddings within a batch:
\begin{equation}
\mathbf{p}_k^c \leftarrow \text{Normalize}\left( (1-\beta_p)\mathbf{p}_k^c +\beta_p \sum_{i \in \mathcal{B}, y_i=c} w_{i,k}^c \mathbf{z}_i \right),
\label{eq:prototype_update_palm}
\end{equation}
where $\mathcal{B}$ denotes the current mini-batch, $w_{i,k}^c$ are the transient soft assignment weights, $\beta_p$ is the EMA momentum parameter, and $\text{Normalize}$ denotes re-projecting the vector onto the unit hypersphere. This gradient-free update mechanism ensures stability during training. The prototype quality score $Q(\mathbf{p}_k^c)$ also adopts the same EMA update mechanism for smooth updating.

%% file: sec/4_Experimental.tex
\section{Experiments}
\label{sec:experiments}

\subsection{Experimental Setup}

\paragraph{Datasets and Training Details}
Our core experiments are conducted on CIFAR-100\cite{CIFAR}, using ResNet-34 as the backbone network. To verify the generalization of our method, we also perform experiments on CIFAR-10 and the larger-scale ImageNet-100 dataset, using corresponding backbones (ResNet-18 and ResNet-50, respectively). Our OOD test sets cover both Far-OOD scenarios (SVHN\cite{c:57}, Textures\cite{c:61}, LSUN\cite{c:59}, iSUN\cite{c:60}, and Places365\cite{places365}) and more challenging Near-OOD scenarios. Unless otherwise specified, all experiments in this section use the CIFAR-100 dataset with the ResNet-34 backbone. Following standard practice, we connect an MLP projection head to the backbone, which embeds features into a 128-dimensional unit hypersphere. All models are trained for 500 epochs using an SGD optimizer with momentum. To ensure fairness, we share the same hyperparameters with the baseline methods wherever possible.
\newline\textbf{OOD Detection Scoring Function} Given that our method aims to learn the true data distribution, we select a widely-used distance-based OOD detection method, the Mahalanobis score \cite{mclachlan1999mahalanobis}. Following standard practice \cite{c:42,sun2022out}, we utilize the feature embeddings from the penultimate layer for distance metric calculation.
\newline\textbf{Evaluation Metrics} We report three commonly used evaluation metrics: (1) The False Positive Rate (FPR) of OOD samples when the True Positive Rate (TPR) of ID samples is 95\%; (2) the Area Under the Receiver Operating Characteristic curve (AUROC); and (3) the Area Under the Precision-Recall curve (AUPR).

\subsection{Inter-Class Prototype Collision}
To validate our hypothesis that prototype collision is a core flaw in baseline methods, we analyzed the trained models.
As shown in Table~\ref{tab:collision}, analysis of the prototype manifold in our reproduced PALM (K=6) model shows its minimum $Q_S$ is $0.000$. Under the threshold $Q_S < 10^{-2}$, we identified 55 pairs of prototypes from different classes colliding, involving 9.17\% of the total prototypes. These collisions primarily occur between classes with high feature similarity, such as leopard $\leftrightarrow$ tiger (sharing similar morphology/texture), rose $\leftrightarrow$ tulip (similar color/shape), and bus $\leftrightarrow$ train (sharing similar linear structures and background contexts). This confirms the existence of discriminatory blind spots caused by the static homogeneity assumption. In contrast, our APM model demonstrates significant improvement: it raises the minimum separation score substantially to $0.013$ and completely eliminates inter-class prototype collisions.

\begin{table}[h] 
\centering
\caption{Analysis of Inter-Class Prototype Collision on \textbf{CIFAR-100 / ResNet-34} models. $Q_S$ denotes the Separation Score.}
\label{tab:collision}
\small 
\begin{tabular}{@{}lccc@{}} 
\toprule
Method & Colliding Prototypes$ \downarrow $ & Min $Q_S$$ \uparrow $ & \% of Total$ \downarrow $ \\ 
\midrule
PALM  & 55 & 0.000 & 9.17 \\
APM   & \textbf{0} & \textbf{0.013} & \textbf{0.00} \\
\bottomrule
\end{tabular}
\end{table}

\subsection{Main Results}

We evaluate the performance of our complete two-stage repair framework APEX on the CIFAR-100 benchmark and our method is compared against a suite of highly competitive baselines, including MSP \cite{c:25}, Vim \cite{wang2022vim}, ODIN \cite{c:43}, Energy \cite{c:37}, VOS \cite{c:29}, CSI \cite{c:40}, SSD+ \cite{c:42}, kNN+ \cite{sun2022out}, NPOS \cite{c:30}, CIDER \cite{ming2022exploit}, PALM \cite{c:35}, and DMPL\cite{c:62}. As shown in Table~\ref{tab:ood_cifar100}, the PALM (K=6) baseline achieves an average FPR of $33.41\%$. 
Our first-stage repair \textbf{APM} significantly enhances the foundational representation quality. The APM model successfully reduces the average FPR by $3.31\%$ to $30.10\%$ (AUROC $92.60\%$) by structurally resolving prototype collisions. This result establishes APM as a superior prototype representation learning foundation.
Further applying our second-stage repair \textbf{PAOS} mechanism (which leads to the full APEX framework), significantly enhances performance across all metrics. The full \textbf{APEX} framework reduces the average FPR further to $\mathbf{29.32\%}$, and the average AUROC reaches $\mathbf{92.91\%}$, achieving new state-of-the-art results. Specifically, APEX achieves the best overall average results and sets new performance records on three out of five Far-OOD datasets (Places365, LSUN, and iSUN), demonstrating the decisive advantage of our two-stage repair framework.

\begin{table*}[!ht]
    \centering
    \caption{OOD detection performance on methods trained on \textbf{labeled CIFAR-100} as ID dataset using backbone network of ResNet-34. ↓ means smaller values are better and ↑ means larger values are better. \textbf{Bold} numbers indicate superior results and \underline{underline} numbers indicate the second-best results.}
    \label{tab:ood_cifar100}
    \small 
    \setlength{\tabcolsep}{5pt}
    \resizebox{\textwidth}{!}{
    \begin{tabular}{@{}l c c c c c c c c c c c c@{}}
        \toprule
        \multirow{3}{*}{Methods} & \multicolumn{10}{c}{OOD Datasets} & \multicolumn{2}{c}{Average} \\
        \cmidrule(lr){2-11} \cmidrule(lr){12-13}
        & \multicolumn{2}{c}{SVHN} & \multicolumn{2}{c}{Places365} & \multicolumn{2}{c}{LSUN} & \multicolumn{2}{c}{iSUN} & \multicolumn{2}{c}{Textures} & \multicolumn{2}{c}{} \\
        & FPR↓ & AUROC↑ & FPR↓ & AUROC↑ & FPR↓ & AUROC↑ & FPR↓ & AUROC↑ & FPR↓ & AUROC↑ & FPR↓ & AUROC↑ \\
        \midrule
        MSP      & 78.89          & 79.80          & 84.38          & 74.21          & 83.47          & 75.28          & 84.61          & 74.51          & 86.51          & 72.53          & 83.57          & 74.62          \\
        Vim      & 73.42          & 84.62          & 85.34          & 69.34          & 86.96          & 69.74          & 85.35          & 73.16          & 74.56          & 76.23          & 81.13          & 74.62          \\
        ODIN     & 70.16          & 84.88          & 82.16          & 75.19          & 76.36          & 80.10          & 79.54          & 79.16          & 85.28          & 75.23          & 78.70          & 78.91          \\
        Energy   & 66.91          & 85.25          & 81.41          & 76.37          & 59.77          & 86.69          & 66.52          & 84.49          & 79.01          & 79.96          & 70.72          & 82.55          \\
        VOS      & 43.24          & 82.80          & 76.85          & 78.63          & 73.61          & 84.69          & 69.65          & 86.32          & 57.57          & 87.31          & 64.18          & 83.95          \\
        CSI      & 44.53          & 92.65          & 79.08          & 76.27          & 75.58          & 83.78          & 76.62          & 84.98          & 61.61          & 86.47          & 67.48          & 84.83          \\
        SSD+     & 31.19          & 94.19          & 77.74          & 79.90          & 79.39          & 85.18          & 80.85          & 84.08          & 66.63          & 86.18          & 67.16          & 85.91          \\
        kNN+     & 39.23          & 92.78          & 80.74          & 77.58          & 48.99          & 89.30          & 74.99          & 82.69          & 57.15          & 88.35          & 60.22          & 86.14          \\
        NPOS     & 10.62          & 97.49          & 67.96          & 78.81          & 20.61          & 92.61          & 35.94          & 88.94          & \textbf{24.92} & 91.35          & 32.01          & 89.84          \\
        CIDER    & 22.95          & 95.17          & 79.81          & 73.59          & 16.19         & 96.32         & 71.96          & 80.54          & 43.94         & 90.42         & 46.97          & 87.21        \\
        PALM     & \textbf{3.03}  & \textbf{99.23} & 67.80          & 82.62          & 10.58          & 97.70          & 41.56          & 91.36          & 44.06          & 91.43          & 33.41          & 92.47          \\
        DMPL    & 16.97  & 96.24 & 73.66          & 79.55          & 13.10          & 96.85        & 40.83          & 89.51          & 35.09         & 90.93         & 35.93          & 90.62          \\
        \midrule 
        \textbf{APM} & $3.41^{\pm 1.04}$ & $99.18^{\pm 0.25}$ & $\underline{64.72}^{\pm 2.00}$ & $\underline{84.62}^{\pm 2.06}$ & $\underline{9.91}^{\pm 2.29}$ & $\underline{97.83}^{\pm 0.77}$ & $\underline{28.79}^{\pm 4.80}$ & $\underline{94.61}^{\pm 1.48}$ & $33.67^{\pm 3.59}$ & $\underline{92.44}^{\pm 0.46}$ & $\underline{30.10}^{\pm 1.29}$ & $\underline{92.60}^{\pm 0.58}$ \\
\textbf{APEX} & $\underline{3.27}^{\pm 1.14}$ & $\underline{99.22}^{\pm 0.22}$ & $\mathbf{63.97}^{\pm 1.79}$ & $\mathbf{84.91}^{\pm 1.89}$ & $\mathbf{9.69}^{\pm 2.36}$ & $\mathbf{97.95}^{\pm 0.83}$ & $\mathbf{28.24}^{\pm 5.20}$ & $\mathbf{94.73}^{\pm 1.32}$ & $\underline{33.31}^{\pm 3.60}$ & $\mathbf{92.69}^{\pm 0.41}$ & $\mathbf{29.32}^{\pm 1.37}$ & $\mathbf{92.91}^{\pm 0.52}$ 
    \\
        \bottomrule
    \end{tabular}
    }
\end{table*}

\begin{table*}[!ht]
    \centering
    \caption{OOD detection performance on methods fine-tuning on ImageNet-100 using pre-trained ResNet-50 models. ↓ means smaller values are better and ↑ means larger values are better. \textbf{Bold} numbers indicate superior results and \underline{underline} numbers indicate the second-best results.}
    \label{tab:ood_imagenet100}
    \small 
    \setlength{\tabcolsep}{9pt} 
    \begin{tabular}{@{}l c c c c c c c c c c@{}}
        \toprule
        \multirow{2}{*}{Methods} & \multicolumn{8}{c}{OOD Datasets} & \multicolumn{2}{c}{Average} \\
        \cmidrule(lr){2-9} \cmidrule(lr){10-11}
        & \multicolumn{2}{c}{SUN} & \multicolumn{2}{c}{Places} & \multicolumn{2}{c}{Textures} & \multicolumn{2}{c}{iNaturalist} & \multicolumn{2}{c}{} \\
        & FPR↓ & AUROC↑ & FPR↓ & AUROC↑ & FPR↓ & AUROC↑ & FPR↓ & AUROC↑ & FPR↓ & AUROC↑ \\
        \midrule
        kNN+     & \textbf{41.85} & 92.25          & 44.41          & 90.26          & 26.60          & 94.22          & 38.54          & 94.15          & 37.85          & 92.72          \\
        CIDER    & 42.26          & 92.84          & 42.81          & 91.39          & 19.31          & 95.44          & 45.49          & 92.83          & 37.47          & 93.12          \\
        PALM     & 43.12          & 92.77          & 42.06          & 91.51          & 17.48          & 95.94          & 32.89          & 94.82          & 33.89          & 93.55          \\
        \midrule 
        \textbf{APM}      & 41.96          & \underline{93.28} & \underline{40.71} & \underline{92.02} & \underline{16.73} & \underline{96.21} & \underline{31.64} & \underline{95.23} & \underline{32.68} & \underline{94.20} \\
        \textbf{APEX} & \underline{41.89} & \textbf{93.31} & \textbf{40.56} & \textbf{92.06} & \textbf{16.59} & \textbf{96.23} & \textbf{31.32} & \textbf{95.27} & \textbf{32.43} & \textbf{94.26} \\
        \bottomrule
    \end{tabular}
\end{table*}

\begin{figure}[!ht]
    \centering \includegraphics[width=0.48\textwidth]{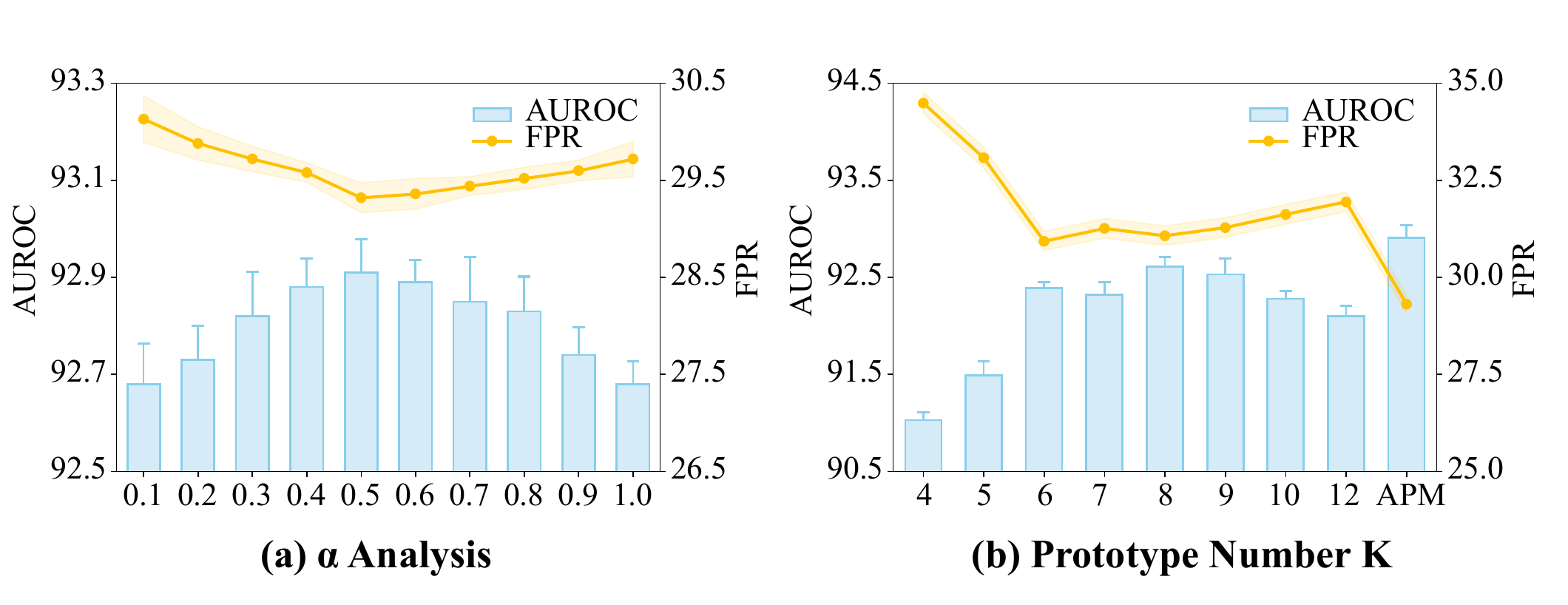} 
        \caption{
       APEX: Experimental analysis of model sensitivity and qualitative results. (a) Analysis of the PAOS calibration strength $\alpha$. (b) Sensitivity to the fixed prototype number $K$.}
    \label{fig:sentivity}
\end{figure}

\subsection{Ablation Study}

\subsubsection{Fixed K vs. APM}

Figure ~\ref{fig:sentivity}(b) compares the performance of different fixed K values against our APM. It is evident that no single fixed K value can match the performance of APM. This proves that adaptively selecting $K_c^*$ based on class complexity is crucial for constructing a high-quality prototype manifold.

\begin{table}[h]
\centering
\caption{Ablation study of PAOS Calibration Components on CIFAR-100. All metrics are averaged (\%).}
\label{tab:qc_maha_ablation_wo}
\small
\sisetup{
    detect-weight = true, 
    mode = text, 
    text-font-command = \textbf
}
\resizebox{\columnwidth}{!}{
\begin{tabular}{@{}l S[table-format=2.2] S[table-format=2.2] S[table-format=2.2]@{}}
\toprule
Calibration  Components & {FPR@95$\downarrow$} & {AUROC$\uparrow$} & {AUPR$\uparrow$} \\ 
\midrule
w/o $Q$ & 30.10 & 92.60 & 89.81 \\
w/o $Q_C$ & 29.62 & 92.86 & 90.32 \\
w/o $Q_S$ & 29.85 & 92.79 & 90.24 \\
Ours & \textbf{29.32} & \textbf{92.91} & \textbf{90.41} \\ 
\bottomrule
\end{tabular}
}
\end{table}

\begin{table}[h]
\centering
\caption{Structural mechanism analysis of $\text{APM}$: $\text{BIC}$ vs. Heuristic $K_c^*$ assignment.}
\label{tab:apm_bic_ablation}
\small
\setlength{\tabcolsep}{3pt}
\resizebox{\columnwidth}{!}{
\begin{tabular}{@{}l ccccc@{}}
\toprule
Method & FPR@95 $\downarrow$ & AUROC $\uparrow$ & Min $Q_S$$ \uparrow $ & Collision $\downarrow$  \\ 
\midrule
Fixed K (PALM) & 33.41 & 92.47 & 0.000 & 55  \\
Random-Uniform & 32.74 & 91.79 & 0.005 & 25  \\
Dirichlet-Noise & 31.89 & 92.04 & 0.007 & 22  \\
Shuffle-of-BIC & 30.81 & 92.41 & 0.008 & 18  \\
\midrule
APM & \textbf{30.10} & \textbf{92.60} & \textbf{0.013} & \textbf{0}  \\
\bottomrule
\end{tabular}
}
\end{table}

\subsubsection{APM Mechanism Analysis: BIC vs. Heuristics}

To prove that APM's success stems from its $\text{BIC}$-driven complexity assignment, not merely from using a dynamic $K_c^*$, we compare it against several heuristics (Table~\ref{tab:apm_bic_ablation}). These methods include $\text{Fixed K=6}$ (PALM), random assignment strategies (e.g., $\text{Random-Uniform}$), and a crucial $\text{Shuffle-of-BIC}$ baseline (which randomly re-allocates the $\text{BIC}$-derived $K_c^*$ values to the wrong classes).
Results in Table~\ref{tab:apm_bic_ablation} validate our approach. Our $\text{APM}$ achieves the best performance ($\text{FPR}=\mathbf{30.10\%}$) and structurally outperforms the $\text{Fixed K=6}$ baseline by dramatically reducing prototype collisions (from 55 to \textbf{0}). Notably, $\text{APM}$ achieves the best separation metric ($\text{Min $Q_S$}=\mathbf{0.013}$). While random heuristics perform poorly, the $\text{Shuffle-of-BIC}$ baseline's failure ($\text{FPR}=30.81\%$) is most revealing: it proves that simply using a dynamic $K_c^*$ set is insufficient. The correct mapping of complexity to class, as determined by $\text{BIC}$, is the cornerstone of $\text{APM}$'s success.

\subsubsection{PAOS Calibration Components}
This ablation study is designed to verify the necessity of each calibration component within PAOS. As presented in Table~\ref{tab:qc_maha_ablation_wo}, the $\text{Ours}$ result achieves the optimal performance ($\text{FPR} = \mathbf{29.32\%}, \text{AUROC} = \mathbf{92.91\%}, \text{AUPR} = \mathbf{92.41\%}$). This demonstrates that the $\text{PAOS}$ mechanism contributes an absolute $\text{FPR}$ reduction of $0.78\%$ atop the $\text{APM}$ baseline, underscoring the value of boundary refinement. Further analysis of component necessity shows that removing either cohesion ($\text{w/o } Q_C$, $\text{FPR}=29.62\%$) or separation ($\text{w/o } Q_S$, $\text{FPR}=29.85\%$) leads to performance degradation. This validates that the dual-quality constraint of $\text{Q}_C$ and $\text{Q}_S$ within the $\text{PAOS}$ mechanism is indispensable for boundary refinement over the $\text{APM}$-optimized feature space, ultimately achieving the superior $\text{APEX}$ performance.

\subsubsection{Sensitivity to Calibration Strength $\alpha$}

As presented in Figure \ref{fig:sentivity}(a), our $\text{PAOS}$ mechanism demonstrates highly stable and superior performance within a broad range of $\alpha$ values, specifically between 0.3 and 0.7, significantly outperforming the boundary conditions of $\alpha=0.1$ and $\alpha=1.0$. Specifically, the model achieves its optimal performance at $\alpha=\mathbf{0.5}$ ($\text{FPR}=\mathbf{29.32\%}, \text{AUROC}=\mathbf{92.91\%}$). This result indicates that the $\text{PAOS}$ mechanism exhibits good \text{robustness} to the hyperparameter $\alpha$, and a medium calibration strength is most effective in enhancing $\text{OOD}$ detection performance.

\begin{table}[ht]
\centering
\caption{Near-OOD detection performance on CIFAR-100. Focused on ImageNet-F, ImageNet-R, and Average Metrics.}
\label{tab:near_ood_single_column_final_2datasets}
\small
\setlength{\tabcolsep}{3.9pt} 
\resizebox{\columnwidth}{!}{
\begin{tabular}{@{}l c c c c c c@{}}
\toprule
\multirow{2}{*}{Methods} & \multicolumn{2}{c}{ImageNet-F} & \multicolumn{2}{c}{ImageNet-R} & \multicolumn{2}{c}{Average} \\
\cmidrule(lr){2-3} \cmidrule(lr){4-5} \cmidrule(lr){6-7}
& FPR$\downarrow$ & AUROC$\uparrow$ & FPR$\downarrow$ & AUROC$\uparrow$ & FPR$\downarrow$ & AUROC$\uparrow$ \\
\midrule
MSP & 86.33 & 70.74 & 86.32 & 72.88 & 87.24 & 72.28 \\
Energy & 78.99 & 76.40 & 80.93 & 80.60 & 84.21 & 74.93 \\
SSD+ & 76.73 & 79.78 & 83.67 & 81.09 & 82.23 & 77.80 \\
KNN+ & 75.52 & 79.95 & 68.49 & 84.91 & 78.27 & 79.04 \\
CIDER & 78.83 & 77.53 & 56.89 & 87.62 & 77.88 & 77.19 \\
PALM & 68.48 & 80.54 & 28.68 & 92.91 & 66.48 & 79.79 \\
\midrule
APM & \underline{65.41} & \underline{81.73} & \underline{27.31} & \underline{93.41} & \underline{63.87} & \underline{80.74} \\
\textbf{APEX} & \textbf{64.89} & \textbf{81.99} & \textbf{26.88} & \textbf{93.63} & \textbf{63.33} & \textbf{81.01} \\
\bottomrule
\end{tabular}
}
\end{table}

\begin{figure}[!t] 
  \centering\includegraphics[width=0.48\textwidth]{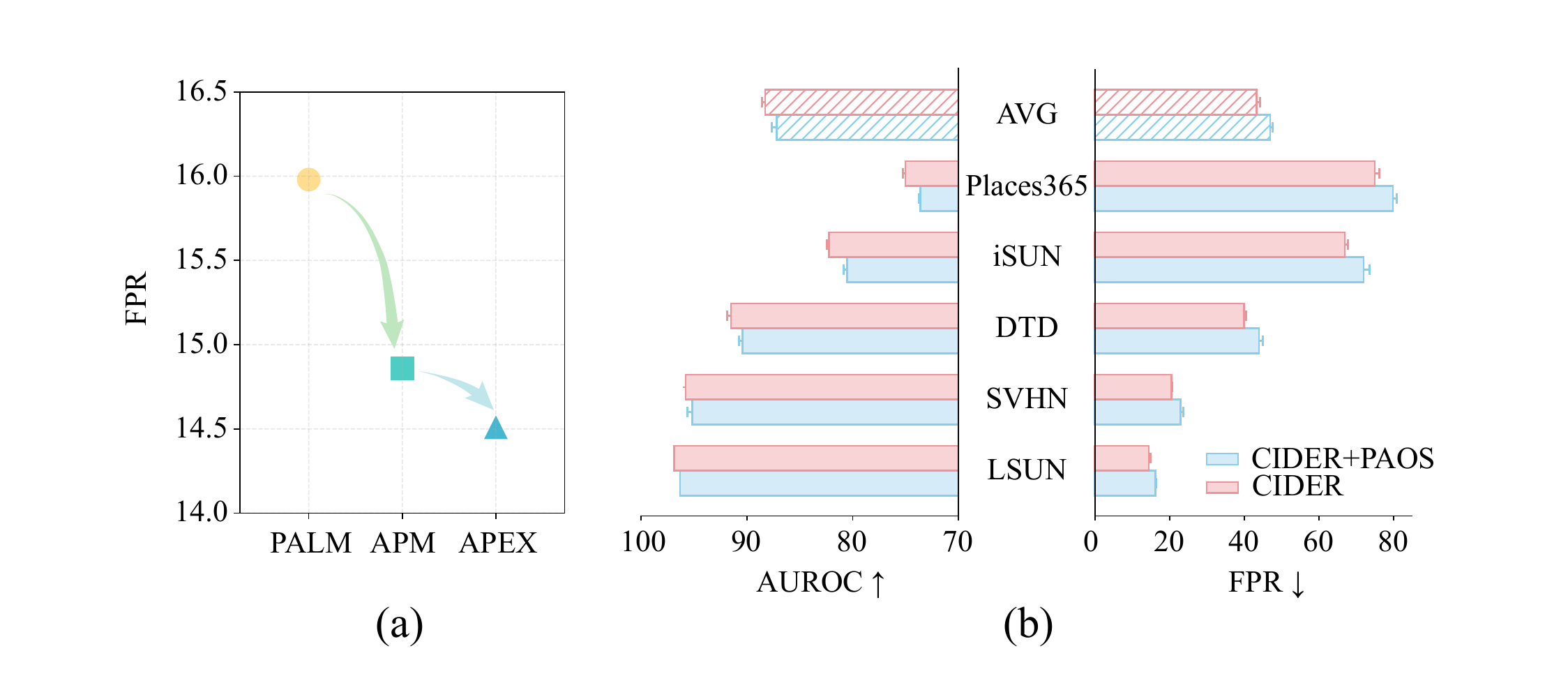}
 
  \caption{
    \textbf{Generalization Performance Analysis of the APM Model and PAOS Module.} 
    \textbf{(a)} Performance comparison of our APM model against baseline methods (e.g., PALM, APEX) when CIFAR-10 serves as the in-distribution dataset, highlighting the superior OOD detection capability of APM.
    \textbf{(b)} Effectiveness validation of the PAOS module as a universal enhancer, illustrating the AUROC performance improvements when integrating it with the existing prototype-based method CIDER on the CIFAR-100 benchmark.
  } 
  \label{fig:cifar10 and cider} 
\end{figure}

\subsection{Generalization Analysis}
\label{sec:generalization}

\subsubsection{Different ID Datasets and Backbones}
We first evaluate APEX on two different scales of ID datasets: CIFAR-10 and ImageNet-100. As demonstrated in Figure ~\ref{fig:cifar10 and cider}(a) and Table~\ref{tab:ood_imagenet100}, our full APEX framework consistently delivers superior average OOD detection performance compared to competitive baselines, including the strong PALM method. For instance, APEX achieves an average AUROC of $\mathbf{97.67\%}$ on CIFAR-10 and $\mathbf{94.26\%}$ on ImageNet-100, affirming that the underlying principles of structural repair and score calibration generalize effectively across varying data complexities and model capacities. This consistent superiority across datasets is particularly significant, as it indicates that the two core flaws we identified are universal. 

\subsubsection{Near-OOD Scenarios}

We further assess our method's robustness on challenging Near-OOD scenarios, where the ID and OOD data distributions are highly similar. Table~\ref{tab:near_ood_single_column_final_2datasets} presents the results focusing on ImageNet-F, ImageNet-R, and Average Metrics. APEX maintains its sota performance, achieving the best average FPR of $\mathbf{63.33\%}$ and AUROC of $\mathbf{81.01\%}$. Specifically, APEX shows a decisive advantage on ImageNet-F and ImageNet-R, demonstrating that its refined prototype manifold is highly effective at distinguishing subtle features shared between ID data and structurally similar OOD samples. The complete results for all Near-OOD datasets are provided in the Appendix.

\subsubsection{PAOS Generalization to Other Prototype Methods}
To demonstrate that our score calibration mechanism, PAOS, is not solely reliant on the APM framework but possesses \text{plug-and-play generalization capability}, we apply PAOS to a sota prototype-based OOD detection method CIDER. We replace CIDER's original scoring function with our PAOS module for a direct comparison. As shown in Figure ~\ref{fig:cifar10 and cider}(b), the integration of PAOS significantly enhances the CIDER baseline, reducing the average FPR by $3.61\%$ (from $46.97\%$ to $\mathbf{43.36\%}$) and improving the average AUROC by $1.08\%$ (from $87.21\%$ to $\mathbf{88.29\%}$). This compelling result validates that PAOS effectively captures fundamental prototype quality information that is applicable across different underlying prototype learning representations, confirming its strong modularity and general utility.

%% file: sec/5_Conclusion.tex
\section{Conclusion}

In this paper, we identify and systematically address two fundamental, overlook flaws universally constraining state-of-the-art prototype-based OOD detection methods: the Static Homogeneity Assumption (leading to catastrophic prototype collision) and the Learning-Inference Disconnect.
To resolve these structural and informational deficiencies, we propose APEX, a novel framework built on a robust Two-Stage Repair process. The framework operationalizes a systematic solution: the APM leverages the MDL principle to eliminate collisions by adaptively allocating resources; and the PAOS mechanism utilizes quantified prototype quality to bridge the learning-inference disconnect.
More fundamentally, our work offers a conceptual insight into the OOD challenge: the problem can be effectively decoupled into a structural challenge (optimizing the manifold, solved by APM) and an inferential challenge (utilizing the posterior knowledge, solved by PAOS). This decoupled framework serves as a practical engineering realization of the \textbf{Information Bottleneck (IB) principle}. In this context, APM manages the adaptive compression aspect, while PAOS ensures the model's decision-making adheres to the sufficiency principle. This principled, efficient approach is key to advancing truly robust representation learning systems.

%% file: main.bib
@String(CVPR= {IEEE Conf. Comput. Vis. Pattern Recog.})

@String(ECCV= {Eur. Conf. Comput. Vis.})

@String(NIPS= {Adv. Neural Inform. Process. Syst.})

@String(AAAI = {AAAI})

@String(CVPR  = {CVPR})

@String(ECCV  = {ECCV})

@String(NIPS  = {NeurIPS})

@article{c:25,
  title={A baseline for detecting misclassified and out-of-distribution examples in neural networks},
  author={Hendrycks, Dan and Gimpel, Kevin},
  journal={arXiv preprint arXiv:1610.02136},
  year={2016}
}

@article{c:26,
  title={Prototypical networks for few-shot learning},
  author={Snell, Jake and Swersky, Kevin and Zemel, Richard},
  journal={Advances in neural information processing systems},
  volume={30},
  year={2017}
}

@article{c:27,
  title={Supervised contrastive learning},
  author={Khosla, Prannay and Teterwak, Piotr and Wang, Chen and Sarna, Aaron and Tian, Yonglong and Isola, Phillip and Maschinot, Aaron and Liu, Ce and Krishnan, Dilip},
  journal={Advances in neural information processing systems},
  volume={33},
  pages={18661--18673},
  year={2020}
}

@inproceedings{wang2022vim,
  title={Vim: Out-of-distribution with virtual-logit matching},
  author={Wang, Haoqi and Li, Zhizhong and Feng, Litong and Zhang, Wayne},
  booktitle={Proceedings of the IEEE/CVF conference on computer vision and pattern recognition},
  pages={4921--4930},
  year={2022}
}

@inproceedings{c:28,
  title={A simple framework for contrastive learning of visual representations},
  author={Chen, Ting and Kornblith, Simon and Norouzi, Mohammad and Hinton, Geoffrey},
  booktitle={International conference on machine learning},
  pages={1597--1607},
  year={2020},
  organization={PmLR}
}

@inproceedings{c:29,
  title={Towards unknown-aware learning with virtual outlier synthesis},
  author={Du, Xuefeng and Wang, Zhaoning and Cai, Mu and Li, Sharon},
  booktitle={International Conference on Learning Representations},
  volume={1},
  number={3},
  pages={5},
  year={2022}
}

@article{c:30,
  title={Non-parametric outlier synthesis},
  author={Tao, Leitian and Du, Xuefeng and Zhu, Xiaojin and Li, Yixuan},
  journal={arXiv preprint arXiv:2303.02966},
  year={2023}
}

@inproceedings{c:31,
  title={Understanding contrastive representation learning through alignment and uniformity on the hypersphere},
  author={Wang, Tongzhou and Isola, Phillip},
  booktitle={International conference on machine learning},
  pages={9929--9939},
  year={2020},
  organization={PMLR}
}

@inproceedings{sun2022out,
  title={Out-of-distribution detection with deep nearest neighbors},
  author={Sun, Yiyou and Ming, Yifei and Zhu, Xiaojin and Li, Yixuan},
  booktitle={International conference on machine learning},
  pages={20827--20840},
  year={2022},
  organization={PMLR}
}

@inproceedings{c:32,
  title={Understanding the behaviour of contrastive loss},
  author={Wang, Feng and Liu, Huaping},
  booktitle={Proceedings of the IEEE/CVF conference on computer vision and pattern recognition},
  pages={2495--2504},
  year={2021}
}

@article{mclachlan1999mahalanobis,
  title={Mahalanobis distance},
  author={McLachlan, Goeffrey J},
  journal={Resonance},
  volume={4},
  number={6},
  pages={20--26},
  year={1999}
}

@article{ming2022exploit,
  title={How to exploit hyperspherical embeddings for out-of-distribution detection?},
  author={Ming, Yifei and Sun, Yiyou and Dia, Ousmane and Li, Yixuan},
  journal={arXiv preprint arXiv:2203.04450},
  year={2022}
}

@article{c:33,
  title={Siren: Shaping representations for detecting out-of-distribution objects},
  author={Du, Xuefeng and Gozum, Gabriel and Ming, Yifei and Li, Yixuan},
  journal={Advances in neural information processing systems},
  volume={35},
  pages={20434--20449},
  year={2022}
}

@article{c:34,
  title={Distributional Prototype Learning for Out-of-distribution Detection},
  author={Peng, Bo and Lu, Jie and Zhang, Yonggang and Zhang, Guangquan and Fang, Zhen},
  booktitle={Proceedings of the 31st ACM SIGKDD Conference on Knowledge Discovery and Data Mining V. 1},
  pages={1104--1114},
  year={2025}
}

@article{c:35,
  title={Learning with mixture of prototypes for out-of-distribution detection},
  author={Lu, Haodong and Gong, Dong and Wang, Shuo and Xue, Jason and Yao, Lina and Moore, Kristen},
  journal={arXiv preprint arXiv:2402.02653},
  year={2024}
}

@article{c:37,
  title={Energy-based out-of-distribution detection},
  author={Liu, Weitang and Wang, Xiaoyun and Owens, John and Li, Yixuan},
  journal={Advances in neural information processing systems},
  volume={33},
  pages={21464--21475},
  year={2020}
}

@article{c:38,
  title={On the importance of gradients for detecting distributional shifts in the wild},
  author={Huang, Rui and Geng, Andrew and Li, Yixuan},
  journal={Advances in Neural Information Processing Systems},
  volume={34},
  pages={677--689},
  year={2021}
}

@article{c:39,
  title={Glow: Generative flow with invertible 1x1 convolutions},
  author={Kingma, Durk P and Dhariwal, Prafulla},
  journal={Advances in neural information processing systems},
  volume={31},
  year={2018}
}

@article{c:40,
  title={Csi: Novelty detection via contrastive learning on distributionally shifted instances},
  author={Tack, Jihoon and Mo, Sangwoo and Jeong, Jongheon and Shin, Jinwoo},
  journal={Advances in neural information processing systems},
  volume={33},
  pages={11839--11852},
  year={2020}
}

@article{c:41,
  title={A simple unified framework for detecting out-of-distribution samples and adversarial attacks},
  author={Lee, Kimin and Lee, Kibok and Lee, Honglak and Shin, Jinwoo},
  journal={Advances in neural information processing systems},
  volume={31},
  year={2018}
}

@article{c:42,
  title={Ssd: A unified framework for self-supervised outlier detection},
  author={Sehwag, Vikash and Chiang, Mung and Mittal, Prateek},
  journal={arXiv preprint arXiv:2103.12051},
  year={2021}
}

@article{c:43,
  title={Enhancing the reliability of out-of-distribution image detection in neural networks},
  author={Liang, Shiyu and Li, Yixuan and Srikant, Rayadurgam},
  journal={arXiv preprint arXiv:1706.02690},
  year={2017}
}

@inproceedings{c:44,
  title={Energy-based open-world uncertainty modeling for confidence calibration},
  author={Wang, Yezhen and Li, Bo and Che, Tong and Zhou, Kaiyang and Liu, Ziwei and Li, Dongsheng},
  booktitle={Proceedings of the IEEE/CVF International Conference on Computer Vision},
  pages={9302--9311},
  year={2021}
}

@article{c:45,
  title={Your classifier is secretly an energy based model and you should treat it like one},
  author={Grathwohl, Will and Wang, Kuan-Chieh and Jacobsen, J{\"o}rn-Henrik and Duvenaud, David and Norouzi, Mohammad and Swersky, Kevin},
  journal={arXiv preprint arXiv:1912.03263},
  year={2019}
}

@article{c:46,
  title={Likelihood ratios for out-of-distribution detection},
  author={Ren, Jie and Liu, Peter J and Fertig, Emily and Snoek, Jasper and Poplin, Ryan and Depristo, Mark and Dillon, Joshua and Lakshminarayanan, Balaji},
  journal={Advances in neural information processing systems},
  volume={32},
  year={2019}
}

@article{c:47,
  title={Generalized out-of-distribution detection: A survey},
  author={Yang, Jingkang and Zhou, Kaiyang and Li, Yixuan and Liu, Ziwei},
  journal={International Journal of Computer Vision},
  volume={132},
  number={12},
  pages={5635--5662},
  year={2024},
  publisher={Springer}
}

@inproceedings{c:57,
  title={Reading digits in natural images with unsupervised feature learning},
  author={Netzer, Yuval and Wang, Tao and Coates, Adam and Bissacco, Alessandro and Wu, Baolin and Ng, Andrew Y and others},
  booktitle={NIPS workshop on deep learning and unsupervised feature learning},
  volume={2011},
  number={5},
  pages={7},
  year={2011},
  organization={Granada}
}

@article{c:59,
  title={Lsun: Construction of a large-scale image dataset using deep learning with humans in the loop},
  author={Yu, Fisher and Seff, Ari and Zhang, Yinda and Song, Shuran and Funkhouser, Thomas and Xiao, Jianxiong},
  journal={arXiv preprint arXiv:1506.03365},
  year={2015}
}

@article{c:60,
  title={Turkergaze: Crowdsourcing saliency with webcam based eye tracking},
  author={Xu, Pingmei and Ehinger, Krista A and Zhang, Yinda and Finkelstein, Adam and Kulkarni, Sanjeev R and Xiao, Jianxiong},
  journal={arXiv preprint arXiv:1504.06755},
  year={2015}
}

@article{c:62,
  title={Enhancing out-of-distribution detection via diversified multi-prototype contrastive learning},
  author={Jia, Yulong and Li, Jiaming and Zhao, Ganlong and Liu, Shuangyin and Sun, Weijun and Lin, Liang and Li, Guanbin},
  journal={Pattern Recognition},
  volume={161},
  pages={111214},
  year={2025},
  publisher={Elsevier}
}

@inproceedings{c:61,
  title={Describing textures in the wild},
  author={Cimpoi, Mircea and Maji, Subhransu and Kokkinos, Iasonas and Mohamed, Sammy and Vedaldi, Andrea},
  booktitle={Proceedings of the IEEE conference on computer vision and pattern recognition},
  pages={3606--3613},
  year={2014}
}

@article{intro:early-method,
  title={Concrete problems in AI safety},
  author={Amodei, Dario and Olah, Chris and Steinhardt, Jacob and Christiano, Paul and Schulman, John and Mané, Dan},
  journal={arXiv preprint arXiv:1606.06565},
  year={2016}
}

@inproceedings{intro:trustworthy-AI-systems,
  title={Why ReLU networks yield high-confidence predictions far away from the training data and how to mitigate the problem},
  author={Hein, Matthias and Andriushchenko, Maksym},
  booktitle={IEEE Conference on Computer Vision and Pattern Recognition (CVPR)},
  pages={41--50},
  year={2019}
}

@inproceedings{intro:discriminative-representations,
  title={Momentum contrast for unsupervised visual representation learning},
  author={He, Kaiming and Fan, Haoqi and Wu, Yuxin and Xie, Saining and Girshick, Ross},
  booktitle={IEEE/CVF Conference on Computer Vision and Pattern Recognition (CVPR)},
  year={2020}
}

@inproceedings{ignores-inter-class-variability,
  title={Robust prototype learning with noisy labels},
  author={Yang, Liu and Jin, Rong and He, Jiebo},
  booktitle={AAAI Conference on Artificial Intelligence},
  year={2018}
}

@inproceedings{ignores-inter-class-variability2,
  title={Adaptive prototypes for few-shot learning},
  author={Liu, Lu and Song, Le and Qin, Yuxuan and Zhang, Bo and Zhang, Guojie},
  booktitle={European Conference on Computer Vision (ECCV)},
  year={2020}
}

@inproceedings{visual-complexity-differences,
  title={Distribution-aware contrastive representation learning for long-tailed recognition},
  author={Kim, Beomsu and Kim, Seong Joon and Kim, Hyunwoo J},
  booktitle={IEEE/CVF Conference on Computer Vision and Pattern Recognition (CVPR)},
  year={2022}
}

@inproceedings{visual-complexity-differences2,
  title={Class-aware prototype learning for imbalanced classification},
  author={Han, Jihwan and Park, Jinheon and Shin, Jinwoo},
  booktitle={International Conference on Machine Learning (ICML)},
  year={2023}
}

@inproceedings{EBMretain-information,
  title={Energy-based out-of-distribution detection in latent space},
  author={Du, Yinpeng and Hu, Han and Zhang, Bo and Dong, Liang and Shi, Yichen and Zhang, Song},
  booktitle={European Conference on Computer Vision (ECCV)},
  year={2022}
}

@techreport{CIFAR,
  title={Learning multiple layers of features from tiny images},
  author={Krizhevsky, Alex},
  year={2009},
  institution={University of Toronto},
  note={Technical report}
}

@inproceedings{places365,
  title={Places: A 10 million Image Database for Scene Recognition},
  author={Zhou, Bolei and Lapedriza, Aude and Khosla, Aditya and Oliva, Antonio and Torralba, Antonio},
  booktitle={IEEE Transactions on Pattern Analysis and Machine Intelligence (TPAMI)},
  volume={40},
  number={6},
  pages={1452--1464},
  year={2018},
  publisher={IEEE}
}

@article{MDL,
  title={Estimating the dimension of a model},
  author={Schwarz, Gideon E},
  journal={The Annals of Statistics},
  volume={6},
  number={2},
  pages={461--464},
  year={1978}
}

@inproceedings{BIC,
  title={Modeling by shortest data description},
  author={Rissanen, Jorma},
  booktitle={Automatica},
  volume={14},
  pages={465--471},
  year={1978}
}

@article{dino,
  title={DINOv2: Learning robust visual features without supervision},
  author={Caron, Mathilde and Touvron, Hugo and Misra, Ishan and Jegou, Hervé and Mairal, Julien},
  journal={arXiv preprint arXiv:2304.07193},
  year={2023}
}

@inproceedings{EMA,
  title={Mean teachers are better role models: Weight-averaged consistency targets improve semi-supervised deep learning results},
  author={Tarvainen, Antti and Valpola, Harri},
  booktitle={Advances in Neural Information Processing Systems (NeurIPS)},
  year={2017}
}

@article{c:63,
  title={Mitigating the modality gap: Few-shot out-of-distribution detection with multi-modal prototypes and image bias estimation},
  author={Wang, Yimu and Riddell, Evelien and Chow, Adrian and Sedwards, Sean and Czarnecki, Krzysztof},
  journal={arXiv preprint arXiv:2502.00662},
  year={2025}
}

@inproceedings{c:64,
  title={Out-of-distribution detection with prototypical outlier proxy},
  author={Gong, Mingrong and Chen, Chaoqi and Sun, Qingqiang and Wang, Yue and Huang, Hui},
  booktitle={Proceedings of the AAAI Conference on Artificial Intelligence},
  volume={39},
  number={16},
  pages={16835--16843},
  year={2025}
}

@article{c:65,
  title={Raising the bar in graph ood generalization: Invariant learning beyond explicit environment modeling},
  author={Shen, Xu and Liu, Yixin and Wang, Yili and Miao, Rui and Dai, Yiwei and Pan, Shirui and Chang, Yi and Wang, Xin},
  journal={arXiv preprint arXiv:2502.10706},
  year={2025}
}
